# Exploring the "Great Unseen" in Medieval Manuscripts: Instance-Level Labeling of Legacy Image Collections with Zero-Shot Models

Christofer Meinecke*, Estelle Guéville†, and David Joseph Wrisley‡

We often think of the visual content of European medieval manuscripts as the cues included by image makers for the metonymic visual identification of figures or historical moments. The term for this is *iconography*: a hand descending from the sky for the creation of the world, Magdalene with her long hair and ointment jar, or Saint Peter with his key (Garnier, 1989; Lipton, 2012). And yet these visual cues can be challenging to identify computationally against the complex visual layouts of manuscripts (Baechler et al, 2010). Earlier phases of art historical database construction and labeling have created a number of starting points for thinking about medieval images. Their labeling methods created generic iconographic metadata for retrieval for close viewing and interpretation, for example, by studying a combination of elements that acquire meaning based on the placement within the text, or the interaction between colors and forms (Baschet, 2008).

In this long paper, we aim to theorize the medieval manuscript page and its contents more holistically. We bring state-of-the-art techniques to bear in segmenting and describing the entire manuscript folio, rather than a singular focus on the historiated initial or prominent visual content linked to traditional iconography. This serves the purpose of creating richer training data for downstream computer vision techniques, namely instance segmentation, and beyond that, multimodal models for medieval-specific visual content. These tasks have significant applicability in a variety of domains, making the specific tasks at hand one that could have significant general impact for scholars of visual culture. For example, manuscripts contain myriad decorative patterns beyond the scope of classic iconography: decorative ascenders or descenders; arches, frames or other forms reminiscent of Gothic architecture (Porcher 1960); introductory miniatures; marginal drolleries, even plant-based motifs such as acanthus, rose or grapevines, all with limited tagging in iconographic databases. Without robust training data, however, these visual clues, especially the grotesque and "edgy" depictions of human-animal hybrids (Camille 1992), are likely to evade computational identification.

In two partial datasets of medieval manuscripts located in French collections we have previously used, such holistic labeling is rare. For Mandragore, zoological annotations have been done using bounding boxes only (Michel 2019). For the rest of Mandragore and the entirety of Initiale, for the included labels, coordinates for localization are missing. As such, these image-level annotations are not well suited for state-of-the-art object detection and image segmentation models. Fine-tuning requires instance-level annotations (bounding boxes or segmentation masks), which are time-consuming to create. Along with the fact that even when there is annotated training data, cultural heritage image data is by definition limited in scale. This situation is a strong argument for zero-shot segmentation models like "Segment Anything" that learn to generalize image segmentation of points of interest without a label vocabulary. The movement from image-level annotation as found in the legacy collections to instance-level annotation is not without its share of complexities (Aouinti et al. 2022; Zhang et al. 2022).

We combine zero-shot segmentation in a custom labeling interface where both legacy annotations can be reused, enhanced or ignored and segmentation masks can be created with few easy interactions. The masks can be labeled using the legacy vocabulary via drag-and-drop, or extended to new labels. Our system uses the SAM2 (Ravi et al. 2024) model for zero-shot segmentation, which can be used without any user input to generate a segmentation of a whole image. It can be used as a starting point by filtering automatically generated polygons that are not of interest. To ensure annotation reliability, the whole process is semi-automatic and all segmentations undergo human validation. On the one hand, the system can be thought of as the logical extension of previous labeling methods, except that precise localization for a given label is now possible, and on the other hand, it allows the extension and enriching of annotations about medieval visual content in a much more granular and inclusive way.

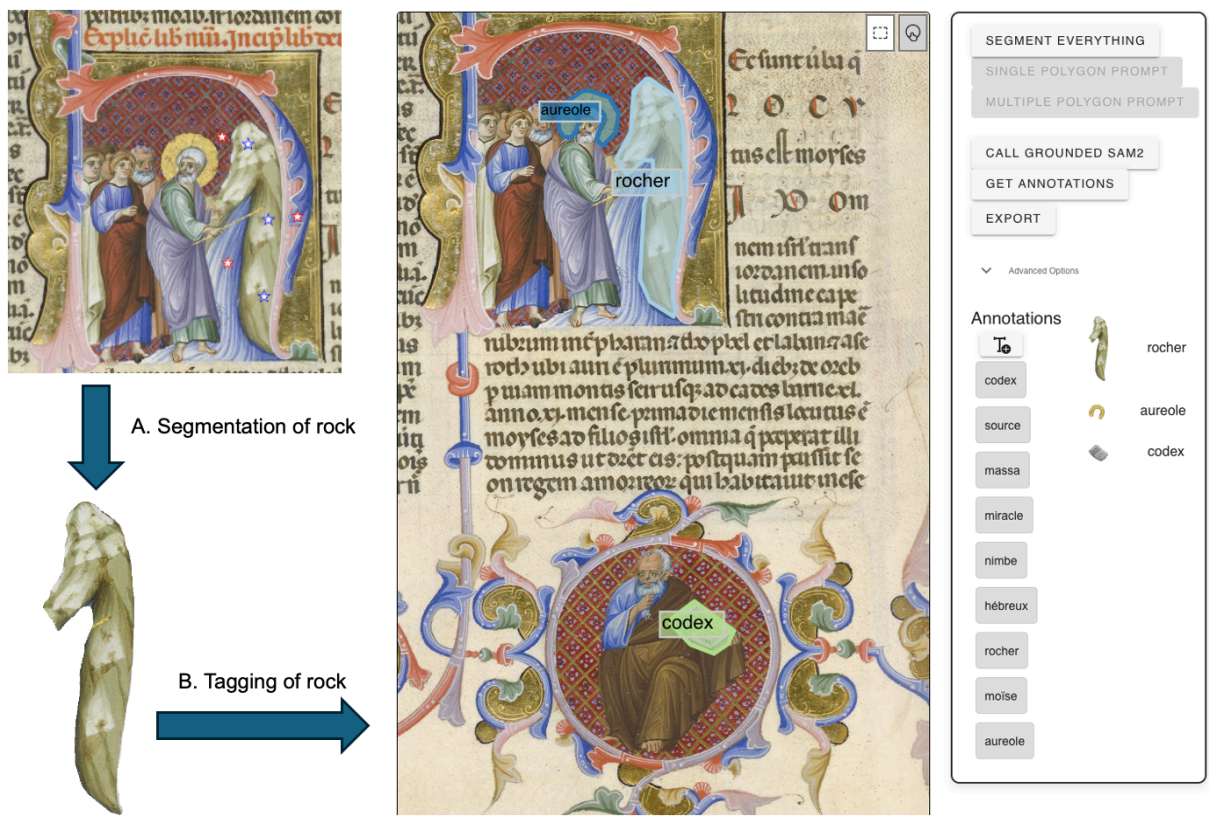

**Figure 1:** An image of our annotation interface. At upper left, point prompts are used to define detailed segmentation masks (blue are positive, and red are negative). The segmentation is generated and labeled (e.g., rocher/rock). Existing labels can be used, as seen in the bas de page medallions (e.g., codex) and new annotations can be created (e.g., auréole/halo). The manuscript depicted here is Paris, Bibliothèque nationale de France, Latin 18, folio 53v.

In terms of interactions, the user can draw a box prompt, defining an area of interest for the segmentation model. Additional control over the segmentation is provided by point prompts, for selecting several points of the image as positive or negative examples in order to segment one polygon of interest (Figure 1). We have found this method most useful to distinguish between parts of an image or to focus on image borders, for example, to the inside of a historiated initial and the letter itself. By setting several positive points, it is also possible to segment several objects of interest at once, although the complex overlapping content of medieval manuscripts renders this difficult. Drag-and-drop of existing annotations allows labeling a polygon, with the polygon consumed the most by the text label in case of overlap. Words can also be moved to other polygons to change the target of the annotation.

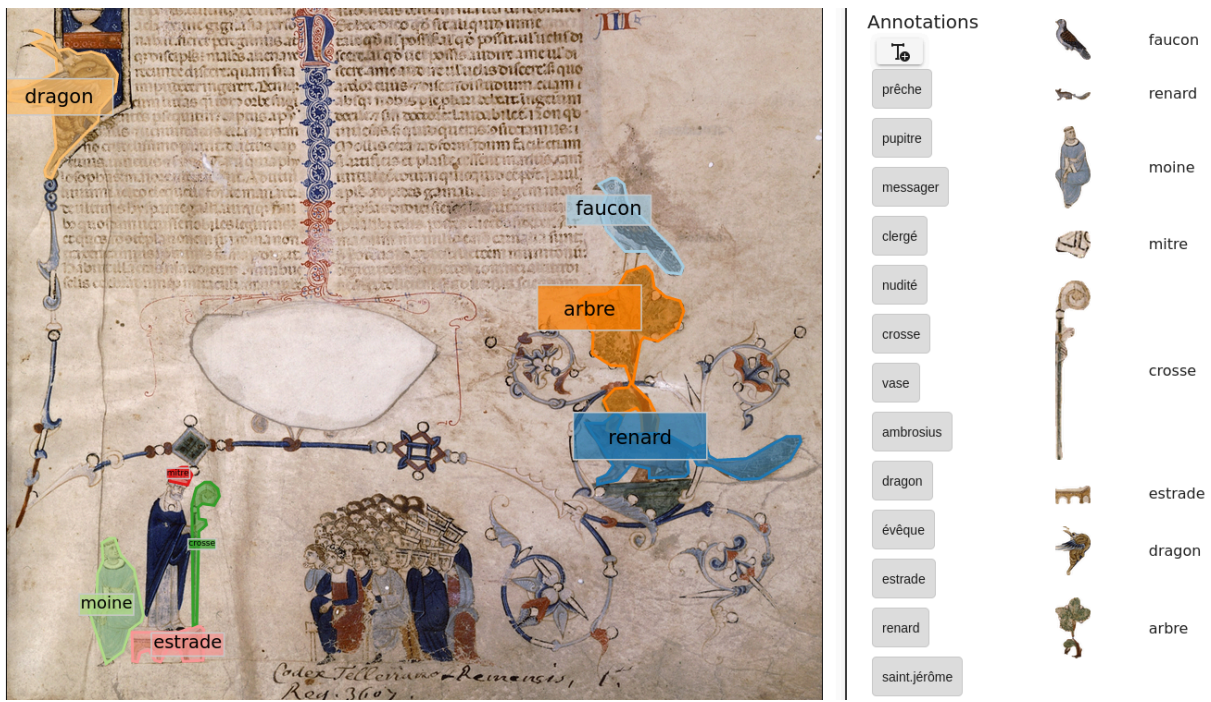

**Figure 2:** An image of our annotation interface where the bounding box tool was used to define detailed (croppable and adjustable) segmentation masks for existing labels (dragon, faucon/falcon, renard/fox, moine/monk, crosse/crozier, estrade/rostrum) in the margins of the manuscript and to create a new annotation (mitre/miter and arbre/tree). Note the visual features defined by the polygon at right. The manuscript depicted here is Paris, Bibliothèque nationale de France, Latin 22, folio 1r.

Previous work has noted the tensions between object- and narrative- or action-centered iconographic vocabulary (Meinecke et al, 2024); here, in order to extend and enrich annotations, we focus on the non-narrative, non-action-centered labels. In doing so, we try to mitigate the difficulties that come with annotation judgments and what is actually automatically identifiable by diffusing the hierarchical distinctions that typically accompany iconographic labeling. We also flatten distinctions usually made in art historical analysis between marginalia, decorative elements and historiated initials in favor of more universal forms. One way to approach such difficulties of iconographic labeling is to think of it as a hierarchy of forms and actions. In Figure 2, objects such as a table, a bench, a sword, a crown, a human being, a crozier or a miter are forms that are easy to identify and label.

However, identifying a bishop requires some interpretation, or, in other words, the combination of three of the objects previously identified to create meaning, or requiring context provided by the manuscript's text, while the identification of the action of preaching requires additional interpretative scenes, such as the presence of a crowd, itself dependent on the presence of lower-level objects (multiple human beings identified as the clergy based on context) (Figure 3). We have not decided how to define such hierarchies at present and our goal is not to evacuate meaning from such objects; instead, we conceive of instance-level segmentation and tagging as a first step to constructing a grammar of low level objects and to facilitate computationally modeled iconography.

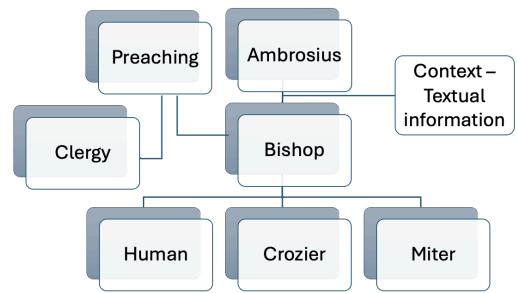

**Figure 3**: An illustration of some contextual information found in manuscripts, showing possible relationships between tags.

Our design in the present anticipates future distant viewing approaches. Arnold and Tilton (2023) have explained how the potentialities of distant viewing are unlocked more fully by domain-specific attention to annotation, and Smits and Wevers argue that zero-shot models can open the way for a multimodal turn in DH research (Smits & Wevers 2023). Previous work has shown how multimodal models can be used to explore visual culture in specific contexts: in collections of photographs from the 1970s (Arnold & Tilton 2024) and for zero-shot, retrieval, and labeling tasks of several visual datasets from the 19th and early 20th century (Smits & Wevers 2023). We are not aware of examples of such models extended to more diverse data, for example, to non-Western or pre-modern visual data.

For that reason, our system is designed to be domain-agnostic, supporting the use of multi-modal models to create image- and instance-level annotations of images for a variety of datasets. For the automatic creation of image-level annotations, we apply RAM++ (Huang et al. 2023 RAM++ paper) and Grounded-SAM for automatic instance segmentation by text prompts (annotations). RAM++ is good at detecting some modern objects in the images but focuses more on the actual domain of the image, excluding domain-specific details such as the rotulus, crozier, or mitre. Grounded-SAM is a combination of Grounding DINO and SAM2, designed to detect and segment anything with text inputs by using Grounding DINO (Liu et al. 2024) to detect bounding boxes based on the given text prompt and by using SAM2 to segment the object inside the detection. Grounding SAM can perform instance segmentation for specific persons, such as a king or a monk, as long as there are only a few people in the image. Despite not being limited to a predefined set of tags, its modern training data nonetheless results in a lack of knowledge about domain-specific vocabulary suitable for medieval manuscripts.

Our annotation system allows us to fine-tune to leverage these multi-modal models for a semi-automated annotation process across a variety of visual datasets, in our case across several legacy collections. Such models have been applied to the multi-modal retrieval of iconclass codes (Banar et al. 2023). Since our objective is to address the visual material of medieval manuscript collections more inclusively than legacy iconographic projects have, RAM++ and Grounding DINO allow us to use existing images, legacy textual annotations, along with new ones and segmentation together with descriptions of the visual content of the images as well as a general textual description of the object for the purposes of fine-tuning to the medieval domain.

Furthermore, the instance-level annotations can be used for a nearest neighbor search across all unlabeled segmented parts to allow for batch labeling. Our ultimate goal is an iterative labeling, recommendation and fine-tuning process that will result in a larger number of instance segmentations that allow a distant viewing of domain-specific classes, and thus, a large-scale analysis of visual patterns of these classes to facilitate new exploration of part of the "great unseen" of these medieval manuscripts.